\documentclass[10pt, a4paper]{article}

\usepackage[final]{lrec2026} 

\title{GPT-NL Public Corpus: A Permissively Licensed, Dutch-First Dataset for LLM Pre-training}

\name{Jesse van Oort$^{*}$ \\ {\bf \large Frank Brinkkemper$^{\dagger}$} \\ {\bf \large Erik de Graaf$^{*}$} \\ {\bf \large Bram Vanroy$^{\ddagger}$} \\ {\bf \large Saskia Lensink$^{*}$}\\}

\address{
$^{*}$TNO, The Netherlands \\
\texttt{\{firstname\}.\{lastname\}@tno.nl} \\
\\
$^{\dagger}$GPT-NL, The Netherlands \\
\texttt{frank.brinkkemper@gpt-nl.nl} \\
\\
$^{\ddagger}$Instituut voor de Nederlandse Taal, The Netherlands \\
KU~Leuven, Belgium \\
\texttt{bram.vanroy@ivdnt.org} \\
\\
}

\abstract{
We present the GPT-NL Public Corpus, the biggest permissively licensed corpus of Dutch language resources. The GPT-NL Public Corpus contains 21 Dutch-only collections totalling 36B preprocessed Dutch tokens not present in any other LLM pretraining corpus. Additionally, the corpus includes roughly 207B English, 232B Code, and 48B German/Danish tokens taken from existing sets which we further curated for compliance. This corpus includes curated data from large existing corpora like Common Corpus and Common Crawl, as well as newly created Dutch-specific collections. Most newly created Dutch collections consist of content collected in collaboration with organisations or synthetically augmented content. All data is collected and evaluated with the aim of facilitating the creation of (commercial) language models that are lawful, useful and non-harmful. All data included in the GPT-NL Public Corpus is sourced from datasets with permissive licensing and is curated and redistributed under a CC-BY license. The full dataset is publicly available on the Hugging Face Hub\footnote{\url{https://huggingface.co/datasets/GPT-NL/GPT-NL_Public_Corpus}}.
 \\ \newline  }

\begin{document}

\maketitleabstract

\section{Introduction}
Training large language models (LLMs) requires billions or even trillions of tokens from unique texts. However, it is not easy to find such a vast amount of curated data while complying with copyright legislation, especially for languages other than English. The result is that it is hard for researchers and developers to work with and train new artificial intelligence (AI) models that perform adequately in such languages. This makes organizations in Dutch-speaking regions dependent on large multilingual models where Dutch is not the main focus during training. Furthermore, with growing attention to copyright \cite{bbc2025anthropic} and transparency \cite{transparency2025whatwewant} in LLM training, and as more of these concerns are being codified into law in the European Union\footnote{\url{https://artificialintelligenceact.eu/}}
, it is important to understand which public data can and should be used for training large language models and to develop tools to detect and create useful data.

The compiled corpus contains all the public, permissively licensed content that was used to train the first version of the GPT-NL model.\footnote{\url{https://gpt-nl.nl/}} The GPT-NL model is trained on both this corpus and the GPT-NL proprietary dataset\footnote{\url{https://huggingface.co/datasets/GPT-NL/Collection-metadata}}, which includes content licensed specifically to GPT-NL and is not made publicly available. This paper only discusses the public portion of the data, which includes selections from existing corpora and newly created collections.

The Related Work (Section~\ref{sec:related-work}) describes other initiatives for creating large-scale corpora with high-quality Dutch data. Section~\ref{sec:guidelines} goes into detail about how we decide which data to include in the corpus based on usefulness, lawfulness and potential harm. Section~\ref{sec:main-creation} describes the actions taken to create and select data for the corpus. Section~\ref{sec:provenance} shows the final selection of curated collections in the Public Corpus. Section~\ref{sec:evaluation} details the evaluation process used to approve data for the Corpus. Finally, the paper ends with a description of the limitations of the corpus in Section~\ref{sec:limitations}.

\section{Related Work}
\label{sec:related-work}
A wide range of Dutch corpora exist that differ in size, domain coverage, and licensing. The largest quantity of Dutch text is available on the web and can be accessed via open projects like Common Crawl \cite{commoncrawl} and the Internet Archive \cite{internet-archive}. These data collections themselves are not filtered for quality (though they respect opt-outs), but derivative projects such as CulturaX \cite{culturax},  FineWeb-2 \cite{penedo2025fineweb2pipelinescale}, and HPLT v3 \cite{hplt} provide a pre-filtered, deduplicated subset of that data. These resources provide massive coverage and are easily accessible for model training, but despite the applied quality and language filters, they may still include noisy documents, depending on the quality of the Dutch-specific filters. More important for us, however, is that their licensing and content provenance are often unclear or underspecified.

The PleIAs team published Common Corpus \cite{langlais2025commoncorpuslargestcollection}, which is a substantial resource of permissively licensed text data. It contains around 2 trillion tokens in total, and the GPT-NL project made use of many of the relevant collections it contains. Dutch is fairly under-represented in that corpus, containing mostly 140 year old content from the Royal Library. There are around 8 billion Dutch tokens in Common Corpus. Section~\ref{sec:main-creation} goes over the subsets of Common Corpus that are included in our dataset.

For curated Dutch corpora, the Instituut voor de Nederlandse Taal (Dutch Language Institute; INT) provides several high-quality datasets via ``Taalmaterialen'', with clear metadata and usage and copyright requirements.\footnote{\url{https://taalmaterialen.ivdnt.org/}} The largest corpus is Lassy Large \cite[Lassy Groot;][]{lassy}, containing approximately 700 million words of syntactically parsed Dutch text. The Corpus Gesproken Nederlands \cite[CGN;][]{oostdijk2000cgn} complements this with about 9 million words of transcribed spoken Dutch, offering broad dialectal and register diversity. Both corpora are widely used in Dutch (computationally) linguistic research and provide clean, high-quality text under controlled licenses, though with restricted commercial reuse. However, subsets of Lassy Groot and CGN are available for commercial use.

Other datasets can be found in the field of machine translation and multilingual modelling. The Dutch Parallel Corpus \cite[DPC;][]{dpc} contains around 10 million copyright-cleared words from professional, journalistic, and administrative sources, provided by professional translators (English, Dutch, French; different translation directions). On the EU-level, resources such as Europarl \cite{koehn-2005-europarl} and DGT-TM \cite{steinberger-etal-2012-dgt} collectively provide millions of words of aligned Dutch text from parliamentary and legal domains.

In short, web-crawl corpora deliver scale but uncertain licensing and quality, while smaller-scale efforts offer clearer legal status of a more controlled quality. For GPT-NL we must bring together existing collections and create additional data following a certain set of guidelines to train large language models based on permissively licensed datasets.

\section{Guidelines for suitable data}
\label{sec:guidelines}
Broadly speaking, datasets must conform to three principles to be included in the GPT-NL Public Corpus: 1. data needs to be \textbf{useful} in furthering the main capabilities of the GPT-NL model; 2. data copyright needs to be \textbf{permissive for both non-commercial and commercial use}; and 3. datasets \textbf{should not contain a large amount of harmful and biased data}. These guidelines are kept in mind whilst selecting and creating data (Sec.~\ref{sec:main-creation}) and the evaluation process of these datasets (Sec.~\ref{sec:evaluation}).

\subsection{The Usefulness Perspective}
\label{sec:usefulness}
An important step in the development of models is thinking about their intended use. One of the largest and well-documented issues plaguing the reliability of LLM technology is the fact that LLMs often hallucinate. Contradictory and noisy data as well as limited knowledge are among the major causes of these hallucinations \cite{bang2025hallulensllmhallucinationbenchmark}. One often used mitigation strategy is to create systems in which the LLM is not the primary source of knowledge, for instance in a retrieval-augmented setup \cite{tonmoy2024comprehensivesurveyhallucinationmitigation}. 

The GPT-NL model is trained from the perspective that LLMs are most useful, not as a technology with an enormous inherent knowledge-base but as a technology that interprets and translates verifiable information (from internal or external databases) to humans or other parts of a software system clearly. To create systems that can be trustworthy and overall useful, we gathered reliable data necessary to train models that can be used in this context.

This means we are not focused on covering every piece of information from sources such as Common Crawl \cite{commoncrawl}. Many developers of large language models do include these vast resources, which helps improve the models' ability to give answers to topics described on the web. However, this approach also raises some important challenges, as described in the next two subsections.

\paragraph{Languages in the Corpus}
The primary use for this corpus is to facilitate training models that understand the Dutch and English language. Therefore, the most useful datasets serve to further model capabilities specifically in these languages. Due to the limited availability of Dutch resources, we decided to supplement the corpus with German and Danish public content, based on evidence that moderate amounts of multilingual data improves performance for low-resource languages, specifically for those that are syntactically similar \cite{chang-etal-2024-multilinguality}. Furthermore, \citet{ma2023trainingstagedoescode} showed that adding code data in the pre-training stage can not only improve a model's coding skills but also improve a model's ``reasoning'' to a certain extent. For this reason, code data is also collected in this corpus. Finally, a second, frequently spoken language in the Netherlands is Frisian, which is also collected to facilitate Frisian understanding by models trained on this corpus. Text in languages other than those mentioned here (Dutch, English, German, Danish, Frisian) is out of scope for the GPT-NL Public Corpus.

\subsection{The Law Perspective}
\label{sec:law}
Our starting point is that data collected to train LLMs should be acquired with permission from copyright holders, either through written consent or through permissive licensing. In the GPT-NL Public Corpus, we focus on data that is useful for both commercial and non-commercial use. In this section, we will discuss the legality of using certain datasets with the intention of creating LLMs. In cases where the law is yet unclear, we stay on the conservative side as we assume that much of the `grey area' data is not intended to be used for LLM training. 

\paragraph{LLM as derivative work}
The discussion on whether LLMs should constitute a derivative work (a transformed version) of their training dataset is yet unresolved and legislation and case law is currently unclear. It can be argued that data, when used to train LLMs, is transformed and/or translated in the model by influencing its weights. In addition, studies \cite{236216,leybzon2024learning} show that LLMs may memorize portions of the training data under certain circumstances, which supports the argument that at least some of the data is included in the LLM in a transformed form. For that reason we take the starting point that the LLM could be considered a derivative work under the Creative Commons (CC) license. This rules out data subject to Non-Derivative (ND) licenses, as such licenses prohibit derivatives of the licensed data. 

\paragraph{Commercial license restrictions}
\label{sec:commercial-license}
As mentioned before, our aggregated dataset should be suitable for both commercial and non-commercial model training. Therefore, data subject to a Non-Commercial (NC) CC license, is not included in the creation of this corpus. For the same reason, we also do not include any data subject to Share-Alike (SA) CC license in the GPT-NL Public Corpus; any LLM trained on this type of data would have to be redistributed under the same permissive licensing terms, given the assumption that an LLM qualifies as a derivative work. For a model that will be commercially used, the requirement to make it available under a Share-Alike Creative Commons license can be undesirable. This leaves CC-0, CC-BY (excl. *-NC or *-SA) and public domain data as permissively licensed text-based data suitable for training data for LLMs that can be commercially exploited.

\paragraph{Code dataset licensing}
\label{sec:code-license}
Open-source code licensing is meant to instruct developers how to use or accredit software in their own work. This is complicated when applied to LLM pretraining data. Due to the vast amount of software licensing possibilities and our commitment to uphold the ethos of such licensing, we decided to focus on code with only the four most commonly used open-source code licensing terms of which the terms are most easily translatable to LLM pretraining licensing terms. This leaves Apache 2.0, MIT, BSD(2/3-Clause), and Unlicense as suitable licenses.

\paragraph{TDM-exception and unclear licensing of web data}
The Common Crawl association scrapes web-content that is not behind a paywall and does not have a digital opt-out (e.g. a robots.txt file) and redistributes the data. However, it is not clear-cut whether using data scraped by the Common Crawl association \cite{commoncrawl} is in compliance with EU law when used in LLM training. The Text and Data Mining (TDM) exception in Article 4 of the EU Directive on Copyright and Related Rights in the Digital Single Market (CDSM Directive)\footnote{\url{https://eur-lex.europa.eu/legal-content/EN/TXT/HTML/?uri=CELEX:32019L0790}} provides that TDM opt-outs for otherwise public content only need to be machine-readable (not only via robots.txt). However, no readily available crawling tools able to identify all possible versions of machine-readable opt-outs with an acceptable error rate. Therefore, solely relying on Common Crawl's adherence to opt-outs in robots.txt files might not be sufficient to respect opt-outs. This is why for the GPT-NL Public Corpus we have not included the full Common Crawl dataset but only portions of it that are subject to permissive license terms (Section~\ref{sec:c5-in-GPC}).

\subsection{The Harm and Bias perspective}
\label{sec:harm-bias}
Bias is one of the biggest issues in LLMs \cite{gallegos-etal-2024-bias}. Many demographic groups are under-represented in any large corpus. As a consequence, models trained on such biased data would also carry over that bias in their output.

Some of the more harmful data that is usually present in datasets used for LLM pre-training is excluded in our corpus. Data from places like Reddit, or even Common Crawl in general, although rich in information about the world, may also be rich with discriminatory, offensive or outdated opinions without thorough filtering. By selecting mostly heavily curated and moderated content, we hope to reduce the amount of harmful bias and offensive content in the dataset. 

\section{Creation of the GPT-NL Public Corpus}
\label{sec:main-creation}
In this section we describe how we have selected data from existing sources, and created new data that fit our aforementioned guidelines for suitable data. This data still passes a curation and evaluation phase (Section~\ref{sec:evaluation}). The following sections are summarized in the overview Table~\ref{tab:collected-datasets}.

\subsection{Collected Data}
\label{sec:collected-data}
Most of the data in the GPT-NL Public Corpus is collected by working together with dataset maintainers in the Netherlands.
\paragraph{Targeted Crawls}
We worked together with the \href{https://openstate.eu/en/}{Openstate Foundation} for targeted crawls. We extracted datasets where we had explicit mention by copyright holders that the data was allowed to be scraped. By getting explicit consent, regulations given in Article 53(1)(c) of the AI Act are immediately met\footnote{\url{https://artificialintelligenceact.eu/article/53/}}. The code used for scraping can be found on the Openstate Github.\footnote{\url{https://github.com/openstate/gpt-nl}}

\paragraph{Digitalization Efforts}
We worked with various archives in the Netherlands to obtain public domain content. All included works are at least 100 years old and often at least several decades older. The content in these collections has been digitized with a variety of Hand-written Text Recognition (HTR) and Optical Character Recognition (OCR) techniques. For the content from ``Het Utrechts Archief'' we performed the digitization process ourselves based on the scans they had available. That dataset was further refined by removing documents with low OCR and language detection indicators.

\subsection{Synthetic Data}
\label{sec:synth-creation}
In the GPT-NL Public Corpus, we restrict ourselves to synthetic data sources where (1) all primary knowledge is already present in a lawful source under the restrictions mentioned before to minimize hallucinations; and (2) there is limited risk of ``leakage'' from pre-trained models, thus maintaining a clean data lineage.

\vspace{0.3em}
We distinguish two categories:
\begin{itemize}
    \item \textit{Type 1:} Synthetic data constructed from permissible and traceable sources, such as machine translation or the transformation of structured datasets.
    \item \textit{Type 2:} Synthetic data generated directly by prompting an LLM to produce new topical text (e.g., encyclopedic paragraphs), where source origins of facts may not be traceable or compliant.
\end{itemize}
Only Type~1 data is included in our final dataset.

\subsubsection{Knowledge Graph Text: Wikidata}
\label{sec:Wikidata}
Due to Wikipedia's CC-BY-SA licensing, the data is excluded from the GPT-NL Public Corpus (see Section~\ref{sec:law}). As a replacement, we utilized Wikidata, licensed under CC-0.\footnote{\url{https://www.wikidata.org/}} Wikidata is a large structured knowledge graph expressing facts as subject--predicate--object triples.

The conversion pipeline is as follows:
\begin{enumerate}
    \item \emph{Static Code Generation:} Custom scripts\footnote{\url{https://github.com/GPT-NL/wikidata_synthetic_data_generator}} mapped Wikidata triples to Dutch sentences. Expressions generated at this stage were repetitive or unnatural.
    \item \emph{LLM Rewriting:} An LLM, Phi-4 \cite[MIT-licensed;][]{abdin2024phi4technicalreport}, was used to paraphrase these outputs for improved readability and variation, resulting in output that, while less repetitive, remained below human parity.
    \item \emph{Filtering:} The data was filtered to:
    \begin{itemize}
        \item Exclude all personal data entries for individuals without a corresponding Wikipedia page, adhering to privacy standards and only retaining notable figures.
        \item Remove entries deemed factually trivial.
    \end{itemize}
\end{enumerate}
\begin{table}[h!]
\centering
\begin{tabular}{l|p{0.55\linewidth}}

\textbf{Triple} & Willem-Alexander -- noble title -- Prince of Orange  \\
\hline
\textbf{Dutch Sentence} & \textit{Willem-Alexander heeft de titel Prins van Oranje.}\\

\end{tabular}
\caption{Example mapping from a Wikidata triple to a raw sentence.}
\label{tab:triple_example}
\end{table}

An example can be seen in Table \ref{tab:triple_example}. This process resulted in a broad extraction of factual knowledge from Wikidata, resulting in the new Wikidata-Synth dataset.

\subsubsection{YouTube Transcript Translation}

YouTube was leveraged as a source of real-world speech data. We used the Youtube-Commons\footnote{\url{https://huggingface.co/datasets/PleIAs/YouTube-Commons}} collection included in Common Corpus and selected transcripts in English and Spanish as these are the most abundant. We excluded those already translated or available in Dutch. Normalization and deduplication were applied and non-linguistic markers (such as ``[laughter]'' or ``[music]'') were removed to the extent feasible. The remaining transcripts were translated into Dutch via the open-source MADLAD-400 machine-translation model \cite[Apache-2.0;][]{kudugunta2023madlad400}. 

\subsection{Filtering web-crawls}
\label{sec:filtered-web}
While massively web-crawled datasets have their risks (Sec.~\ref{sec:guidelines}), it is hard to ignore the potential of the enormous amount of relevant data available on the web. With the creation of C5 in collaboration with Instituut voor de Nederlandse Taal and KU~Leuven, we introduce a tool that can help annotate licensing metadata on web-crawled content. C5 was further scrutinized and manually filtered to be included in the GPT-NL Public Corpus.

\subsubsection{The Common Crawl Creative Commons Corpus (C5)}
\label{sec:c5}
While the web contains a large amount of data that is not explicitly licensed for use, there are also websites that explicitly mark their content's license. A common example is Wikipedia, which has a footer on all of its pages, explicitly stating that the contents on the page are licensed under ShareAlike restrictions (which are against the Commercial license restrictions, Sec. ~\ref{sec:commercial-license}).

Unlike earlier web corpora such as C4Corpus \cite{habernal-etal-2016-c4corpus} or dolma-cccc \cite{soldaini-etal-2024-dolma}\footnote{\url{https://huggingface.co/datasets/nkandpa2/cccc_all_domains}}, C5 does not rely purely on string matching to detect ``creativecommons.org'' URLs with elaborate regular expressions. Instead, C5 performs full HTML parsing and structural license extraction: each HTML document is parsed into a DOM-like representation, and candidate license links or metadata (e.g. \texttt{<meta name="license">}, JSON-LD, \texttt{<a>} links, \texttt{<link>} tags) are located and annotated. From these candidate licenses, C5 retains metadata about where the license was found (in head, body or footer; link tag vs. generic <a>, etc.), whether multiple licenses in a document conflict, and confidence ordering (e.g. meta tag in head is ranked higher than an arbitrary link in body). This structural approach allows careful post-filtering to minimize false positives (for example, when a discussion of a CC license appears in text or is specific to an image on the page) -- a control that regex-only methods generally cannot provide. 

In this first release, seven Common Crawl crawls were processed. Only these languages were retained \cite[as identified by GlotLid;][]{kargaran-etal-2023-glotlid}: Afrikaans, German, English, French, West-Frisian, Italian, Dutch and Spanish. The total number of documents and number of tokens are given in Table~\ref{tab:c5_counts}.

\begin{table}[ht]
\centering
\label{tab:c5_counts}
\begin{tabular}{lrr}
\hline
\textbf{Language} & \textbf{Documents} & \textbf{Tokens} \\
Afrikaans & 312{,}262 & 358{,}873{,}448 \\
German & 9{,}530{,}746 & 11{,}362{,}859{,}534 \\
English & 92{,}635{,}372 & 87{,}537{,}859{,}958 \\
French & 9{,}234{,}900 & 12{,}366{,}480{,}025 \\
Frisian & 230{,}910 & 197{,}430{,}774 \\
Italian & 10{,}734{,}597 & 11{,}913{,}669{,}333 \\
Dutch & 2{,}827{,}636 & 2{,}757{,}074{,}705 \\
Spanish & 22{,}226{,}944 & 22{,}515{,}709{,}432 \\
\hline
\textbf{Total} & \textbf{147{,}733{,}367} & \textbf{149{,}009{,}957{,}209} \\
\hline
\end{tabular}
\caption{Document and token counts C5}
\end{table}

Crucially, the dataset is not deduplicated. By releasing the dataset in full, we enable others to apply their own custom filtering and deduplication configurations.

The Common Crawl Creative Commons Corpus is publicly available on the Hugging Face Hub\footnote{\url{https://huggingface.co/datasets/BramVanroy/CommonCrawl-CreativeCommons}} and is fully reproducible with the open-source data processing pipeline which is on Github\footnote{\url{https://github.com/BramVanroy/CommonCrawl-CreativeCommons}}.

\subsubsection{C5 in GPT-NL Public Corpus}
\label{sec:c5-in-GPC}
For the GPT-NL Public Corpus, we opt for strict filtering, as we want certainty about the licensing: zero false positives, which in this context has the meaning of pages wrongly classified as having a permissive open license in line with the guidelines (Sec. \ref{sec:guidelines}). This may happen, for instance, when the C5 pipeline detects a CC-BY license on the footer of the page, but which effectively on refers to the copyright that rests on the icons used on the website, e.g. ``Icons used on this website were licensed under a CC-BY license''. The version of C5 in the GPT-NL Public Corpus is therefore filtered down from the full version.

The filtering steps for the C5 collection within the GPT-NL public corpus are as follows:
\begin{enumerate}
    \item \emph{Filter on domain level:} Group documents from C5 per domain and remove any that do not have CC-0, CC-mark or CC-BY licensing.
    \item \emph{Filter on license position:} Only keep documents where the license is found in the header or footer. Most permissively identified data was annotated based on these license positions. Furthermore, these positions provide licensing information for entire domains compared to single documents, making manual verification feasible.
    \item \emph{Verify and filter domains:} Sort domains by (unverified) permissive tokens per domain. Then verify whether the annotation was done correctly by manually checking for the top domains in token-quantity (>250k words per domain). Only keep domains with manually verified licensing.
\end{enumerate}

In total the filtered C5 set results in 37M words. By publishing C5 in its entirety separately from this GPT-NL Public Corpus, anyone can decide on their own filtering procedure in order to create a version suitable for their needs.  

\subsection{Selected Data}
Common Corpus has 50 collections, varying from public domain (old) data, CC-BY sets (such as Youtube-Commons), CC-BY-SA sets (such as Wikipedia), some open-source software and more. Although not the main contribution of this paper, we have selected and curated collections from Common Corpus that were in line with the suitable data guidelines, and we republish them as part of the GPT-NL Public Corpus.

\section{Evaluation}
\label{sec:evaluation}
\begin{figure*}[ht]
    \centering
    \includegraphics[width=\textwidth]{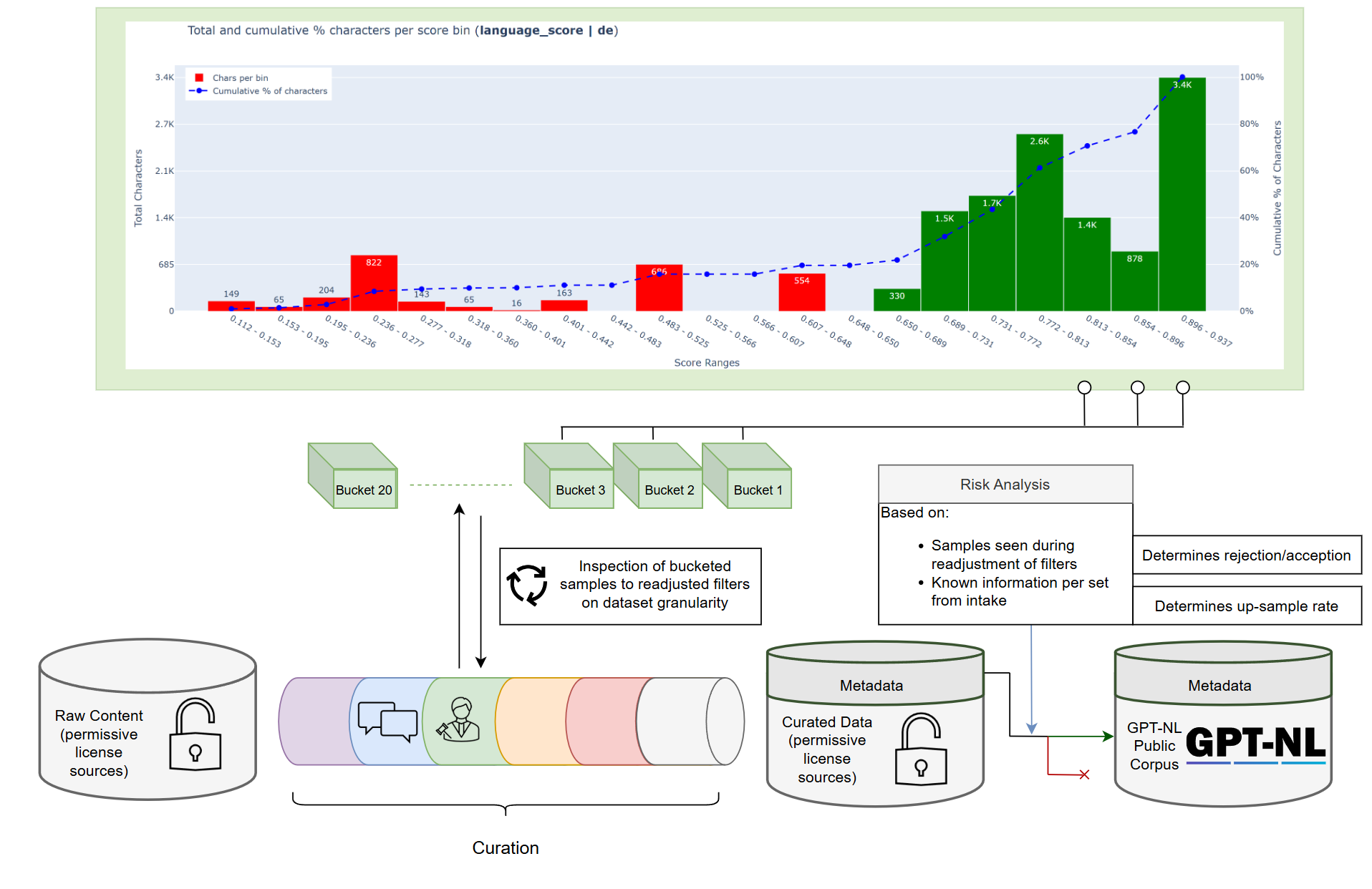}
    \caption{Pipeline transforming raw content into curated and evaluated GPT-NL collections. Here visualized with example manual inspection of the scores in the language dimension (language\_score \textbar{} de).}
    \label{fig:eval-pipeline}
\end{figure*}

\subsection{Curation Process}
\label{sec:curation}
Although not the focus of this paper, it is important to emphasize that all data mentioned in Sec. ~\ref{sec:main-creation} has gone through the GPT-NL Curation Pipeline. This pipeline can be found on the website of GPT-NL \footnote{\url{https://github.com/GPT-NL/data-curation-pipeline}}.

\subsection{Evaluation Process}
For each subset, we gathered as much metadata as possible during the collection and creation of the GPT-NL Public Corpus, for example by talking to the creators or collectors of the original sets. We call this the ``global information''. This information is necessary to evaluate whether a dataset adheres to the guidelines (Sec.~\ref{sec:guidelines}) and gives us the possibility of providing more documentation for every dataset in the GPT-NL Public Corpus. The information collected is included as metadata in the publication of the corpus.

However, this global information is often not enough to definitively approve a dataset. It is important to inspect individual datasets before we can approve them to be part of the GPT-NL Public Corpus. Figure~\ref{fig:eval-pipeline} visualizes the full process of how the raw data is made ready for LLM training. First, a representative sample is taken from the full dataset and sent through the first three phases of the GPT-NL curation pipeline for an `evaluation run'. Visualized from left to right in Figure~\ref{fig:eval-pipeline}, these phases are normalization (purple), language detection (blue) and heuristic filtering (green). 

\subsubsection{Inspection during curation}
The intermediate results of the evaluation run are sent to the data evaluation team. The data in the sample is ranked and visualized according to the scores calculated in the evaluation run across 5 language dimensions (Dutch, English, German, Frisian, and Danish) and 11 heuristic filter dimensions (e.g. duplicate lines, minimum characters). Examples of sentences between different set score-thresholds (or `buckets') are manually inspected. This benefits the final dataset in two ways:
\begin{itemize}
    \item It provides a better idea of the make-up of the dataset. Manually looking at a lot of examples, specifically those that border on getting filtered out by default filter thresholds in one of the different dimensions, gives an idea of the worst data that passed the filters in the curated dataset for any given set. This information can be used during the risk analysis later on.
    \item It is not desirable to treat all data that goes into the curation pipeline in the same way. Some data might be better or worse based on what that data represents. For example, the Dutch language filter might attribute dialects as `less Dutch', attributing a lower Dutch language score and therefore filtering out the data more easily compared to non-dialect Dutch, which would be undesirable to maximally cover Dutch varieties. Going through this process helps fine-tuning the curation parameters on a dataset-granularity.
\end{itemize}

After this process of fine-tuning each filter threshold, the revised filters are applied to the full set, and three post-processing steps are applied: personal data removal, harmful language removal and deduplication. The final dataset can then be subjected to a risk analysis.

\subsubsection{Risk Analysis}
Before the dataset can be officially approved for the GPT-NL Public Corpus, the data goes through a risk analysis. This analysis serves two functions:
\begin{itemize}
    \item Based on the global information collected and the data inspected during curation, a dataset can still be fully rejected during this stage if it is deemed in conflict with the guidelines set up in Section~\ref{sec:guidelines}. At this stage the emphasis lies on the harm and bias perspective. Based on information gathered, we rationalize how including the dataset could potentially harm the quality of a model that is training on it, on basis of harmful language in the data, the potential risk of included personal information, or simply because of the low quality of the data.
    \item For pre-training the GPT-NL model, we have certain token number targets in mind. To achieve those numbers during training, we change the distribution of the input data, upsampling some portions of the dataset more than others. One of the core indicators of how much a single dataset will be up-sampled is the outcome of the risk analysis. Datasets deemed at high risk to have some negative influence on model outputs can generally be expected to be upsampled less than dataset deemed low risk. This is a mitigation to reduce the potential of generating harmful output by models trained on the corpus.
\end{itemize}

The official distribution of the GPT-NL Public Corpus includes metadata about potential risks of individual datasets and the risk level that got attributed to them.

\section{Dataset Overview}
\label{sec:provenance}
In Table \ref{tab:collected-datasets} the total number of tokens, using the GPT-NL tokenizer \footnote{\url{https://github.com/GPT-NL}}, per collection are stated. Collections with the 'CC-' prefix are sourced from Common Corpus. Collections with the 'Synth' suffix have been altered (see Sec. \ref{sec:synth-creation}) before curation. Note that all content, including the Common Corpus collections have been curated according to the GPT-NL curation process (Sec. ~\ref{sec:curation}).

\begin{table*}[ht]
\centering
\begin{tabular}{|p{3.2cm}|p{7cm}|p{1.3cm}|p{1.3cm}|p{1.3cm}|}
\hline
\textbf{GPT-NL Corpus} & \textbf{Content / Source Description} & \textbf{NL (B)} & \textbf{EN (B)} & \textbf{Oth (B)} \\
\hline
\multicolumn{5}{|l|}{\textbf{Collected Data (GPT-NL Curated)}} \\[0.6em] \hline
Openraadsinformatie                  & Municipal council documentation         & 14.1   & 0.02   & 0.01 \\ 
Officiële bekendmakingen              & Government announcements               & 2.8    & 0.01   & -    \\ 
Woogle                               & Open Dutch government documents         & 2.6    & 0.14   & -    \\ 
Koninklijke Bibliotheek               & Public domain Dutch texts               & 2.4    & -      & -    \\ 
De Rechtspraak                        & Judicial cases                          & 2.3    & -      & -    \\
Tweede Kamer                          & Dutch parliamentary documents           & 1.3    & -      & -    \\ 
Nationaal Archief                     & Dutch archive                           & 1.1    & -      & -    \\ 
European Parliament                   & Multilingual EU documents               & 0.05   & 0.03   & 0.02 \\ 
Utrechts Archief                      & Dutch archive                           & 0.2    & -      & -    \\ 
Noord-Hollands Archief                & Dutch archive                           & 0.2    & -      & -    \\ 
Zeeuws Archief                        & Dutch archive                           & 0.2    & -      & -    \\ 
Dienst Publiek en Communicatie        & Dutch public communication docs         & 0.07   & -      & -    \\ 
Wikiwijs                              & Dutch school content                    & 0.03   & -      & -    \\ 
PBL                                   & Planbureau Leefomgeving docs            & 0.02   & -      & -    \\ 
Naturalis                             & Biological publications                 & 0.02   & 0.12   & 0.01 \\ 
DANS-KNAW                             & Dutch archaeology descriptions          & 0.02   & -      & -    \\ 
Auditdienst Rijk                      & Dutch audit publications                & 0.005  & -      & -    \\[0.6em] \hline

\multicolumn{5}{|l|}{\textbf{Selected Data (GPT-NL Curated)}} \\[0.6em] \hline
Belgian Journal                       & Belgian company bylaws (Flemish focus)  & 0.7    & -      & -    \\ 
CC-Eurovoc                               & Multilingual EU vocabulary              & 0.6    & 1.3    & 16   \\
MultiEURLEX                                & EU law texts                            & 0.09   & 0.08   & 0.3  \\
CC-OpenAlex                              & Academic corpus                         & 0.06   & 48     & 0.4  \\
CC-English-PD                            & English public domain texts             & 0.02   & 132    & 0.5  \\ 
American-stories                      & U.S. public domain literature           & -      & 17.6   & -    \\ 
CC-Loc-PD-Books                          & Library of Congress public domain books & -      & 7.5    & -    \\
CC-Github-Code                           & Open code data                          & -      & -      & 232  \\ 
CC-German-PD                             & German public domain texts              & -      & 0.3    & 31   \\
[0.6em] \hline

\multicolumn{5}{|l|}{\textbf{Synthetic Data (GPT-NL Curated)}} \\[0.6em] \hline
CC-Youtube-Commons-Synth                      & Public domain YouTube transcripts translated    & 6.2    & -      & -    \\ 
Wikidata-Synth                  & Wikidata triples converted to running text     & 1.3    & -      & -    \\[0.6em] \hline

\multicolumn{5}{|l|}{\textbf{Filtered web-crawl Data (GPT-NL Curated)}} \\[0.6em] \hline
C5 Filtered                           & Web content                             & 0.04   & -      & -    \\ 
\hline

\multicolumn{2}{|p{10.2cm}|}{\textbf{Total}} & 36.425 & 207.1  & 280 (48 excl. code) \\[0.6em] \hline
\end{tabular}
\caption{Overview of datasets in the GPT-NL Public Corpus}
\label{tab:collected-datasets}
\end{table*}

\paragraph{Collected Data}
Most of the new data included in the GPT-NL Public Corpus was collected in collaboration with governmental, research, archival and judicial organisations. The governmental content includes public content from various parts of the Dutch and Belgian government. For example: meeting documents from the councils of municipalities, provinces, and the House of Representatives (Openraadsinformatie, Tweede Kamer), law and other state announcements (Officiële bekendmakingen, Belgian Journal), and government publications (PBL, Dienst Publiek en Communicatie). Archival content is collected in collaboration with many of the Dutch archives and its content is diverse, as it contains books, newspapers, and letters from many Dutch writers. In terms of judicial content, publicly available court rulings from de Rechtspraak are included as a collection. This includes more than 750,000 documents from the 21st century. Then, some smaller collections are included from a variety of sources. The Wikiwijs collection contains teaching materials from high schools, DANS-KNAW shared their descriptions of archaeological findings in the Netherlands, and Naturalis shared their publications as collections for the GPT-NL Public Corpus. 

\paragraph{Selected Data}
The GPT-NL Public Corpus has included and curated six Common Corpus collections and three additional collections that are in line with Sec.~\ref{sec:guidelines}. The most recent content is sourced from CC-OpenAlex (open science publications), CC-Eurovoc (EU documentation), and MultiEURLEX (European laws) collections. To obtain sufficient English and German content for pre-training, we opted to use some of the larger public domain collections from Common Corpus: CC-English-PD,CC-German-PD and CC-Loc-PD-Books. The American Stories \cite{dell2023american}, MultiEURLEX \cite{chalkidis2021multieurlexmultilingualmultilabel} and Belgian Journal \cite{belgian-journal} collections are included as supplementary content in line with the guidelines. The code included in the GPT-NL Public Corpus is sourced solely from the CC-GitHub-Code collection from Common Corpus.

\paragraph{Synthetic Data}
The synthesized collections are created and included to add more data with a lot of author diversity (YouTube-Commons-Synthesized) and inherent public knowledge (Wikidata-Synth).

\paragraph{Filtered web-crawl Data}
A small set of web content with verified permissive licensing terms.

\section{Limitations}
\label{sec:limitations}
The motivation and funding for creating this corpus is to provide sufficient pre-training content for GPT-NL. Therefore, the corpus is not assumed to be all encompassing in terms of permissively licensed Dutch content.
For the GPT-NL project, it was required during the data acquisition process to invest time in datasets that could be used for the creation of a model that could also be used for commercial activities. It was decided to be legally conservative in terms of licensing and therefore we only include public domain, CC-0 and CC-BY online data sources. This means that commonly used sets like Wikipedia are not included due to it being licensed under CC-BY-SA.
This Corpus is specifically very useful for training models that understand the Dutch and English language. However, the corpus does not have a large amount of (recent) information about the world in comparison to other web-based corpora like FineWeb-2. For the training of the GPT-NL model, this is addressed by also including proprietary licensed datasets specifically for the project, which are not part of the public corpus introduced in this paper. A total of approximately 25 billion additional Dutch tokens from agreements with the national consortium of news publishers in the Netherlands, and other partners are included in the full pre-training corpus of GPT-NL. We will only release metadata about these datasets, but not the data itself.
Some collections are included despite having a questionable result in terms of the harm and bias requirement (sec: \ref{sec:harm-bias}). For example: in the archival collections there is a lot of harmful content, but these collections have a lot of historical context about Dutch culture, so the usefulness perspective made us decide to include them. We recommend inspecting this content before deciding whether you include it in your corpus and when deciding on your data sampling strategy.
In a few of the collections, e.g. `de Rechtspraak', and `Officiële Bekendmakingen', there are many documents that follow similar structures. The court rulings for example use the same section titles for many of the documents, and use similar wording to begin and end sections.
There has been no cross collection deduplication of documents. This is especially relevant for the collections of `Openraadsinformatie', `Woogle', and `Officiële Bekendmakingen', which might include the same document multiple times intended for different audiences.
The automated replacement of personal identifiable information was manually verified with a sample of 100 documents. In this manual verification, some false positives were found, meaning the information of something non-personal was replaced. No false negatives were found in the sampled documents, but due to the limited number of documents, there could still be some non-replaced personal information in the GPT-NL Public Corpus.

\section{Acknowledgments}
The creation of the GPT-NL public corpus is done for the GPT-NL project which is funded by the Ministry of Economic Affairs of the Netherlands and executed by TNO in collaboration with SURF and The Netherlands Forensic Institute (NFI).
The creation of C5 (Section~\ref{sec:c5}) was supported by the VSC (Flemish Supercomputer Center), funded by the Research Foundation Flanders (FWO) and the Flemish Government – department EWI.
For the collected datasets Openraadsinfromatie, Officiële bekendmakingen, Koninklijke Bibliotheek, Naturalis, PBL, and European Parliament the Open State Foundation completed the sourcing of the content.
We want to thank the proactive stance of the people from the content departments of the Rijksoverheid, VNG, Woogle, Koninklijke Bibliotheek, De Rechtspraak, Tweede Kamer, Nationaal Archief, Utrechts Archief, Noord-Hollands Archief, Zeeuws Archief, Wikiwijs, PBL, Naturalis, and DANS-KNAW.
We also thank Simone van Bruggen and Thomas van Osch (SURF), and Martino Mensio, Julia García Fernández, and Erik Vullings (TNO) for their contributions to the OCR and synthetic-data components of this dataset. We thank Morrison Foerster and Leah Griffioen- van Putten (TNO) for legal guidance. Additionally, we thank the entire GPT‑NL team, especially the data-acquisition team (in particular Daan Vos (TNO)) and the data-curation team for their efforts in creating the GPT-NL Public Corpus.
\section{Bibliographical References}\label{sec:reference}

\bibliographystyle{GPTNL_Public_Corpus-natbib}
\bibliography{GPTNL_Public_Corpus}

\label{lr:ref}
\bibliographystylelanguageresource{GPTNL_Public_Corpus-natbib}
\bibliographylanguageresource{languageresource}

\end{document}